\documentclass{article}

\usepackage{arxiv}
\date{}
\renewcommand{\today}{}

\usepackage[utf8]{inputenc} 
\usepackage[T1]{fontenc}    
\usepackage{hyperref}       
\usepackage{url}            
\usepackage{booktabs}       
\usepackage{amsfonts}       
\usepackage{nicefrac}       
\usepackage{microtype}      
\usepackage{amsmath}        
\usepackage{lipsum}
\usepackage{graphicx}
\graphicspath{ {./images/} }
\usepackage{float}
\usepackage{natbib}
\usepackage{xcolor}
\usepackage{wrapfig}
\definecolor{Highlight}{RGB}{0,150,0}
\newcommand{\highlight}[1]{\textcolor{Highlight}{{\textbf{#1}}}}
\usepackage{algorithm}
\usepackage{algpseudocode}   
\usepackage{caption}   

\definecolor{hotpurple}{RGB}{180, 80, 200}

\title{TokenTrim: Inference-Time Token Pruning for Autoregressive Long Video Generation}

\author{
 Ariel Shaulov \\
  School of Computer Science\\
  Tel Aviv University, Israel\\
  \texttt{arielshaulov@mail.tau.ac.il} \\
   \And
 Eitan Shaar \\
  Independent Researcher\\
  \texttt{shaarei@biu.ac.il} \\
  \And
 Amit Edenzon \\
  School of Mathematics\\
  Bar-Ilan University, Israel\\
  \texttt{amit.edenzon@live.biu.ac.il} \\
  \And
 Lior Wolf \\
  School of Computer Science\\
  Tel Aviv University, Israel\\
  \texttt{wolf@cs.tau.ac.il} \\
}

\begin{document}
\maketitle
\vspace{-30pt}
\begin{center}
{\large \href{https://arielshaulov.github.io/TokenTrim/}{\textcolor{hotpurple}{\textbf{TokenTrim Project Page}}}}
\end{center}

\begin{figure}[H]
  \centering
  \includegraphics[width=\linewidth]{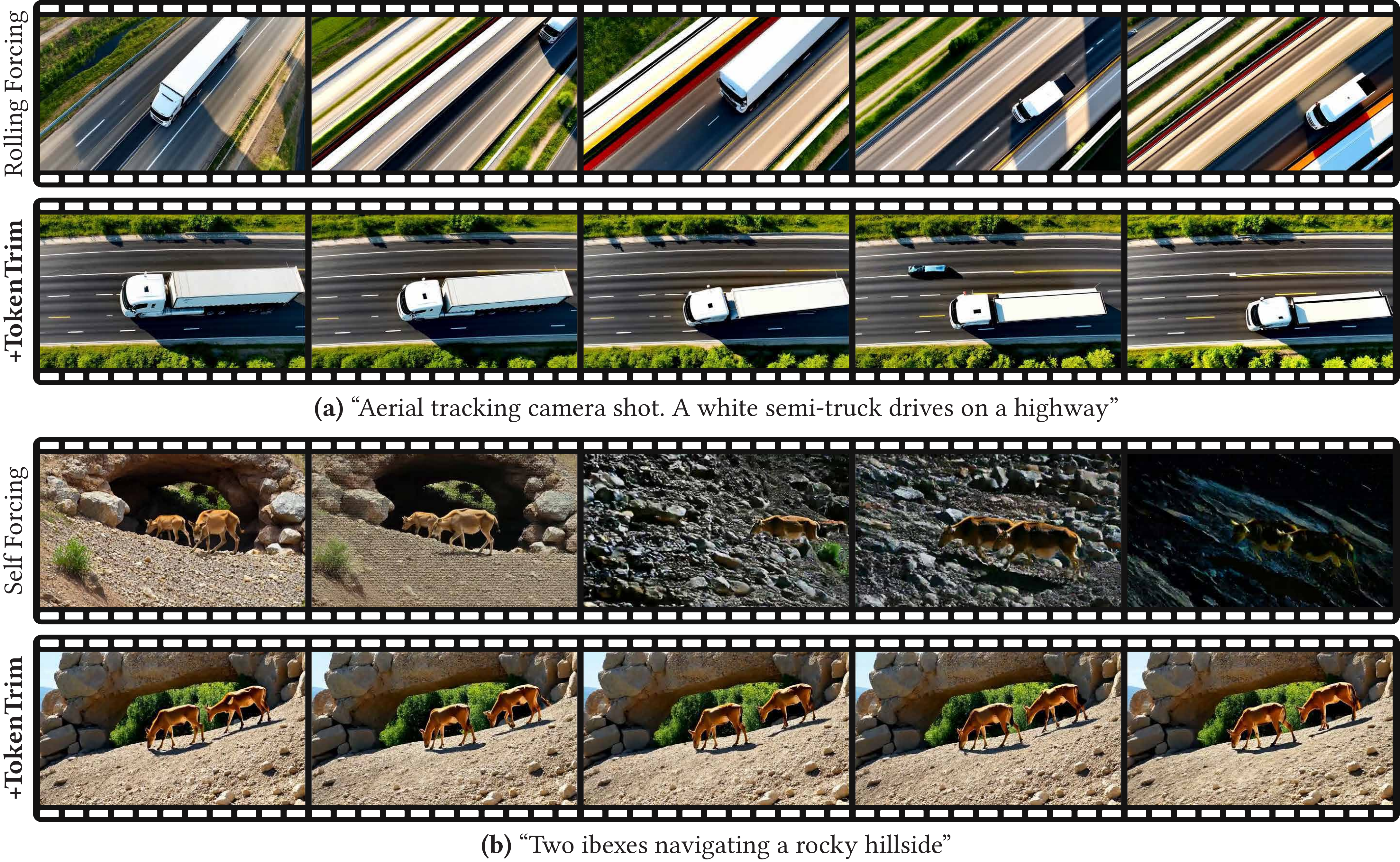}
  \caption{Text-to-video results before and after applying TokenTrim on Rolling Forcing~\cite{liu2025rolling} and Self Forcing~\cite{huang2025self}. 
  }
  \label{fig:teaser}
\end{figure}

\begin{abstract}
    Auto-regressive video generation enables long video synthesis by iteratively conditioning each new batch of frames on previously generated content. However, recent work has shown that such pipelines suffer from severe temporal drift, where errors accumulate and amplify over long horizons. We hypothesize that this drift does not primarily stem from insufficient model capacity, but rather from inference-time error propagation. Specifically, we contend that drift arises from the uncontrolled reuse of corrupted latent conditioning tokens during auto-regressive inference. To correct this accumulation of errors, we propose a simple, inference-time method that mitigates temporal drift by identifying and removing unstable latent tokens before they are reused for conditioning. For this purpose, we define unstable tokens as latent tokens whose representations deviate significantly from those of the previously generated batch, indicating potential corruption or semantic drift. By explicitly removing corrupted latent tokens from the auto-regressive context, rather than modifying entire spatial regions or model parameters, our method prevents unreliable latent information from influencing future generation steps. As a result, it significantly improves long-horizon temporal consistency without modifying the model architecture, training procedure, or leaving latent space.
\end{abstract}


\section{Introduction}
Video diffusion models have rapidly advanced, enabling text-conditioned generation of high-quality videos with realistic appearance and expressive motion~\cite{ho2022vdm,ho2022imagenvideo,singer2022makeavideo,wang2023modelscope,blattmann2023svd,lumiere2024,openai2024sora}.
Despite this progress, generating \emph{long} videos that remain temporally coherent is still a major open challenge~\cite{zhang2024chunkwise,huang2025self,liu2025rolling,longlive2025}.
The dominant recipe for long-horizon synthesis is \emph{chunk-wise autoregressive generation}: the model produces a short clip, then extends the sequence by generating the next clip while conditioning on latent representations of previously generated frames~\cite{zhang2024chunkwise,gao2024ca2vdm,huang2025self,liu2025rolling}. While effective for extending duration, this autoregressive loop often exhibits \emph{temporal drift} (error accumulation): small artifacts and inconsistencies introduced early can compound across chunks, leading to identity changes, structural degradation, and loss of global coherence~\cite{huang2025self,liu2025rolling,longlive2025}.

Recent work has proposed to mitigate drift by strengthening long-range conditioning and stabilizing the autoregressive mechanism, e.g., through temporal key-value (KV) caching~\cite{huang2025self,gao2024ca2vdm}, anchor/sink tokens~\cite{liu2025rolling,longlive2025}, and cache refresh or management strategies during training~\cite{longlive2025,gao2024ca2vdm}.
However, long rollouts remain brittle at {inference time}~\cite{liu2025rolling,zhang2024chunkwise}.
A key reason is that the conditioning context itself degrades: once a region of the latent state becomes corrupted, it is repeatedly reused in subsequent steps and can dominate attention, effectively propagating errors forward~\cite{liu2025rolling}.
This suggests that temporal drift is not only a modeling or data issue, but also an \emph{inference-time information propagation} problem: the system lacks a mechanism to assess which cached latent tokens are trustworthy.

In this work, we introduce \textbf{TokenTrim}, an inference-time method for identifying unstable latent tokens before reuse and removing them from the conditioning context. 
TokenTrim operates entirely in latent space, requires no architectural changes or retraining, and adds only negligible overhead. At each autoregressive step, our method estimates per-token drift by comparing a compact summary of the previous chunk's latent state to the current chunk at its first denoising iteration. Tokens with high drift are hard-pruned from the cached context, preventing corrupted regions from influencing future generations.

TokenTrim is compatible with autoregressive video diffusion frameworks that rely on self-attention with KV caching, including Self Forcing~\cite{huang2025self} and Rolling Forcing~\cite{liu2025rolling}. 
It is also compatible with inference-time regularization techniques such as FlowMo~\cite{shaulov2025flowmo}, which can be applied independently as an additional motion-consistency term. 
By pruning high-drift cached tokens and retaining only reliable context for reuse, TokenTrim suppresses error amplification across autoregressive steps and improves long-horizon temporal consistency. 
Our results show that \emph{controlling the conditioning context at inference time} alone can substantially reduce temporal drift, highlighting an underexplored mechanism for long video generation~\cite{liu2025rolling,zhang2024chunkwise}.

In summary, we propose \textbf{TokenTrim}, a fully inference-time drift detection and hard-pruning mechanism that operates in latent space. We further show that combining motion-stabilized initialization with selective latent pruning significantly improves long-video coherence (see Fig. ~\ref{fig:teaser}), without retraining or modifying the underlying model.

\begin{figure*}[t]
  \centering
  \includegraphics[width=\textwidth]{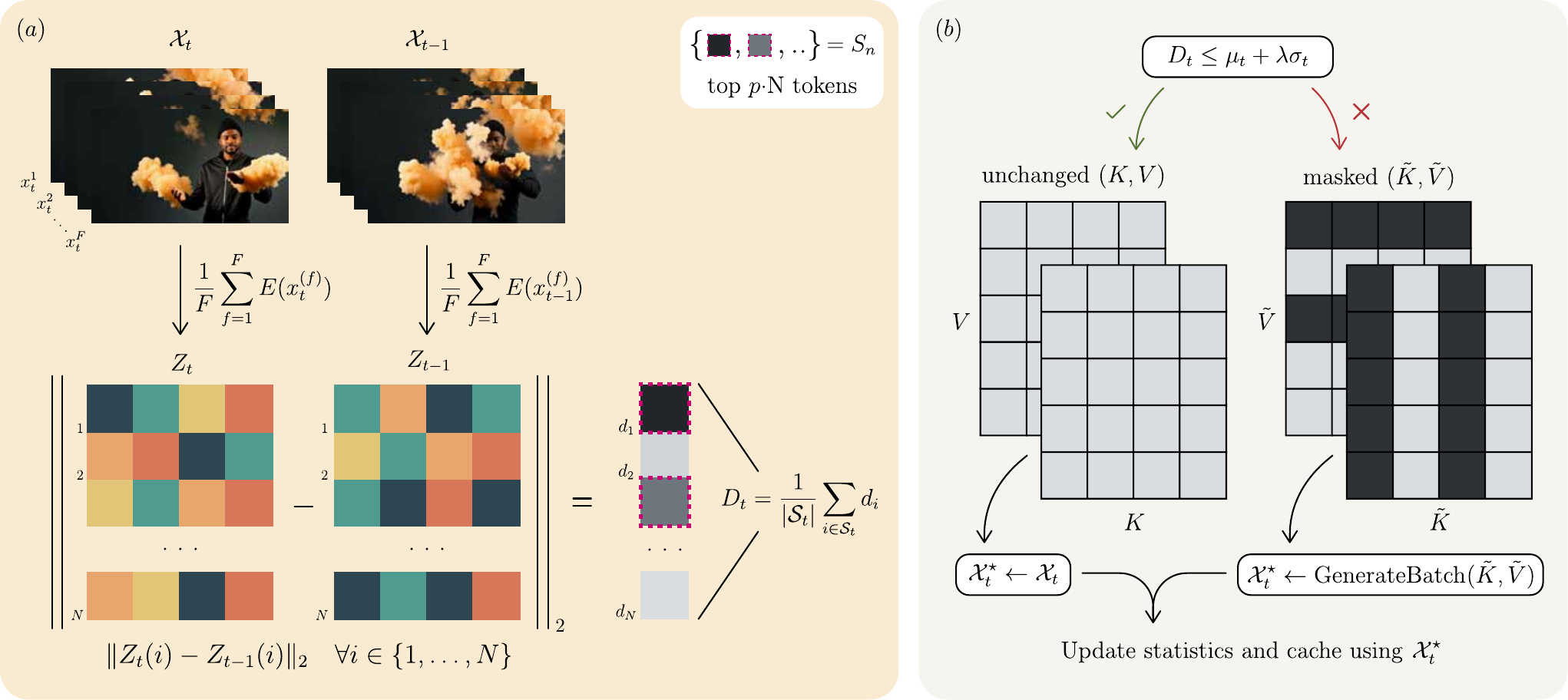}
  \caption{TokenTrim overview at autoregressive step $t$.
\textbf{(a)} Given the candidate batch $\mathcal{X}_t$ and the previous batch $\mathcal{X}_{t-1}$, we encode each frame and form latent summaries
$Z_t$ and $Z_{t-1}$ by averaging latents over the $F$ frames in each batch.
We compute per-token drift $d_i=\lVert Z_t(i)-Z_{t-1}(i)\rVert_2$ and select the top-$pN$ largest drifts to form the unstable set $S_t$, from which we compute the drift severity $D_t$. \textbf{(b)} We compare $D_t$ to the adaptive threshold $\mu_t+\lambda\sigma_t$. If $D_t \le \mu_t+\lambda\sigma_t$, the KV cache $(K,V)$ is left unchanged and the batch is accepted. Otherwise, we mask the selected token positions in the temporal KV cache to obtain $(\tilde{K},\tilde{V})$  and regenerate the current batch conditioned on the pruned cache. Running statistics and the cache are updated using the accepted batch $\mathcal{X}_t^{\star}$. }
  \label{fig:method_figure}
\end{figure*}

\begin{figure*}[p]
  \centering
  \includegraphics[width=0.9\textwidth]{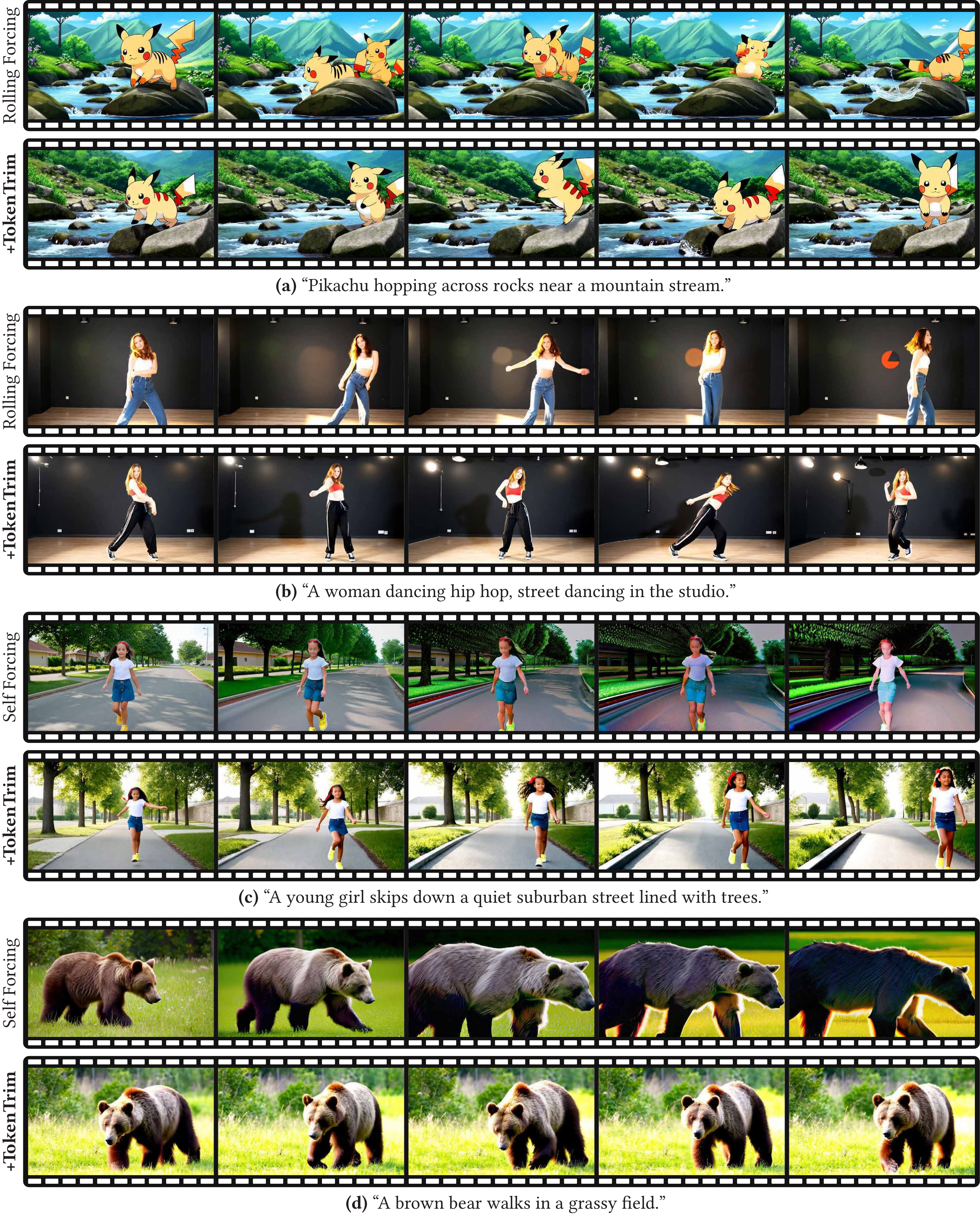}
  \caption{\textbf{Qualitative results.} Text-to-video results before and after applying TokenTrim on Rolling Forcing~\cite{liu2025rolling} and Self Forcing~\cite{huang2025self}. TokenTrim mitigates degradation over time, e.g., color shifts (c - background and girl, d - background and bear), artifacts (b - light lens flare) and unnatural motion (a - Pikatchu). For additional qualitative results see App. ~\ref{app:additional_qualitative_self} and App. ~\ref{app:additional_qualitative_rolling}.}
  \label{fig:qualitative}
\end{figure*}

\section{Related Work}
\subsection{Auto-regressive Long Video Generation} The transition from short-clip synthesis \cite{singer2022makeavideo, ho2022imagenvideo, blattmann2023svd, guo2024animatediff, lumiere2024, chen2024videocrafter2, ge2024preserve} to long-horizon video generation \cite{openai2024sora, kuaishou2024kling, cogvideox2024, lin2024opensora, hacohen2024ltx, pasini2024cam, yin2023nuwaxl, wang2025loong, brooks2022long} has necessitated a shift from holistic spatiotemporal modeling to causal, auto-regressive architectures that factorize generation into sequential chunks. Early approaches like NUWA-XL \cite{yin2023nuwaxl} employed a hierarchical "coarse-to-fine" strategy, generating global key frames before filling local gaps. However, as this precludes real-time streaming, contemporary state-of-the-art methods favor continuous auto-regressive modeling, although often struggling with the ``train-test discrepancy'' or ``exposure bias''. Self Forcing \cite{huang2025self} addresses this by training on self-generated roll-outs with stochastic gradient truncation, ensuring the model learns to recover from its own inference artifacts. To bypass the computational cost of bidirectional attention in streaming contexts, CausVid \cite{yin2025CausVid} distills a bidirectional teacher into a block-causal student, enabling efficient frame-by-frame generation. Addressing the stability of infinite generation, Rolling Forcing \cite{liu2025rolling} introduces "relaxed causality" via rolling-window joint denoising and an attention sink mechanism to anchor identity. Similarly, LongLive \cite{longlive2025} incorporates a KV re-cache mechanism to handle interactive prompt changes without breaking temporal continuity. These training-based methods often require substantial computational resources to realign the model's internal distribution with the auto-regressive task. {However, these training-based methods often suffer from substantial computational costs to realign the model's distribution. More critically, they remain inherently vulnerable to snowballing errors. Even with extensive tuning, minor artifacts inevitably persist in the context and compound into severe temporal drift. In contrast, our approach overcomes this by actively detecting and pruning unstable latent tokens, offering the distinct additional benefit of {suppressing} error propagation at the source to sustain long video consistency without the need for expensive retraining.}

\subsection{Inference-time guidance}

Inference-time guidance offers a training-free paradigm to enhance generation quality by modifying the sampling dynamics or context. To improve temporal consistency, FreeInit \cite{wu2024freeinit} and FreeLong \cite{lu2024freelong} leverage spectral analysis, iteratively refining the initial noise distribution to align low-frequency components with the training manifold, thereby stabilizing global structure. For motion coherence, \citet{shaulov2025flowmo} introduces a latent-optimization approach, calculating the patch-wise temporal variance across generated frames and applying gradient updates to minimize incoherent motion trajectories. Other methods rely on explicit conditioning anchors. For instance ConsistI2V \cite{ren2024consisti2v} modifies spatiotemporal attention to attend to the high-frequency details of the first frame, preventing identity degradation. More recently, focus has shifted to managing the Key-Value cache during inference. While TeaCache \cite{liu2024teacache} and TaoCache \cite{fan2025taocache} prune tokens primarily for acceleration. This demonstrates that selective context management can serve as a powerful form of guidance, a direction our work advances by pruning based on latent instability.

\subsection{Error-Accumulation In Long Video Generation}
A pervasive failure mode in auto-regressive video generation is temporal drift ~\cite{pasini2024cam}, where minor errors in early frames accumulate linearly or exponentially, leading to "error-accumulation" and semantic collapse. Theoretical analysis attributes this to exposure bias, i.e., the distributional shift between the ground truth history seen during training and the imperfect self-generated history at inference~\cite{liu2025rolling, huang2025self, guo2025end}. Empirical studies typically quantify this drift via metrics such as Fréchet Video Distance (FVD) \cite{wang2025loong, zhou2023magicvideo, cogvideox2024} which measures The overall visual quality and temporal coherence of the generated videos or the $\Delta$ Quality Drift introduced in Rolling Forcing \cite{liu2025rolling}. To mitigate drift, architectures like StreamingT2V \cite{henschel2025streamingt2v} employ specialized long-term memory modules (Appearance Preservation Module) to reinject features from an anchor frame. Bagger \cite{po2025bagger} proposes a self-supervised training scheme that aggregates backward trajectories to correct drift. However, these methods often rely on rigid anchoring or extensive retraining. Recent findings in FreeLong \cite{lu2024freelong} suggest that drift manifests non-uniformly across frequency bands, with high-frequency details degrading faster than low-frequency structure. {However, rather than implicitly balancing frequency domains, we propose to explicitly intercept error accumulation at the token level.}

\section{Preliminaries: Self-Attention in Autoregressive Text-to-Video Models}

Modern text-to-video (T2V) generators~\cite{villegas2022phenaki, huang2025self, gao2024ca2vdm, longlive2025} often produce long videos by iteratively synthesizing the output in \emph{chunks} (batches of frames), while conditioning each new chunk on the prompt and on representations of previously generated content~\cite{zhang2024chunkwise,gao2024ca2vdm,longlive2025}. 
Many such systems employ a diffusion backbone for within-chunk generation, often implemented with diffusion transformers (DiTs)~\cite{peebles2023dit,cogvideox2024,lumiere2024}, together with a recurrent/iterative mechanism across chunks to extend temporal horizon~\cite{zhang2024chunkwise}.

At the core of these models are latent spatiotemporal representations and self-attention. Each generated chunk is represented as a set of latent tokens (e.g., patchified latent features over space and time). When generating subsequent chunks, the model conditions on text features as well as latent tokens from earlier chunks, enabling long-range temporal dependencies~\cite{cogvideox2024,lumiere2024}.

\paragraph{\textbf{Self-Attention and Temporal Key-Value Caching}} Self-attention~\cite{vaswani2017attention} allows each token in the current chunk to aggregate information from other tokens in its context, spanning spatial (within-frame) and temporal (across-frame) dimensions. In video DiTs, this mechanism is repeatedly applied inside the denoising network across diffusion steps~\cite{peebles2023dit,cogvideox2024}. 

To support long-horizon context efficiently, many chunk-wise T2V architectures use a temporal key-value (KV) cache that stores key/value projections from previously generated chunks, so current queries can attend to both current tokens and cached tokens without recomputing past projections~\cite{gao2024ca2vdm,longlive2025}. 
Related inference-time caching methods for video diffusion further accelerate generation by selectively reusing computations across denoising steps~\cite{kahatapitiya2024adacache,liu2024teacache}.

Formally, let $\mathbf{Q}$ be the queries for the current tokens, and let $(\mathbf{K}_{\text{cache}}, \mathbf{V}_{\text{cache}})$ denote cached keys/values from earlier chunks. The attention operation is
\begin{equation}
\mathrm{Attention}(\mathbf{Q},\mathbf{K},\mathbf{V}) =
\mathrm{softmax}\!\left(\frac{\mathbf{Q}\mathbf{K}^\top}{\sqrt{d}}\right)\mathbf{V},
\end{equation}
with
\begin{equation}
\mathbf{K} = [\mathbf{K}_{\text{curr}};\mathbf{K}_{\text{cache}}], \quad
\mathbf{V} = [\mathbf{V}_{\text{curr}};\mathbf{V}_{\text{cache}}],
\end{equation}
where $[\cdot;\cdot]$ denotes concatenation along the token dimension and $d$ is the attention head dimension.

\paragraph{\textbf{Temporal Drift}} While temporal KV caching enables long-range dependencies, it can also amplify \emph{temporal drift}: imperfections in earlier latent tokens may be repeatedly attended to and propagated as the cache grows. Over long rollouts, this can accumulate and manifest as identity changes, structural inconsistency, or degraded motion coherence~\cite{zhang2024chunkwise,longlive2025}. 
This mechanism motivates inference-time interventions that control how information is retrieved from the cache and how errors propagate in recurrent T2V generation.

\begin{figure*}[t]
  \centering
  \includegraphics[width=0.91\textwidth]{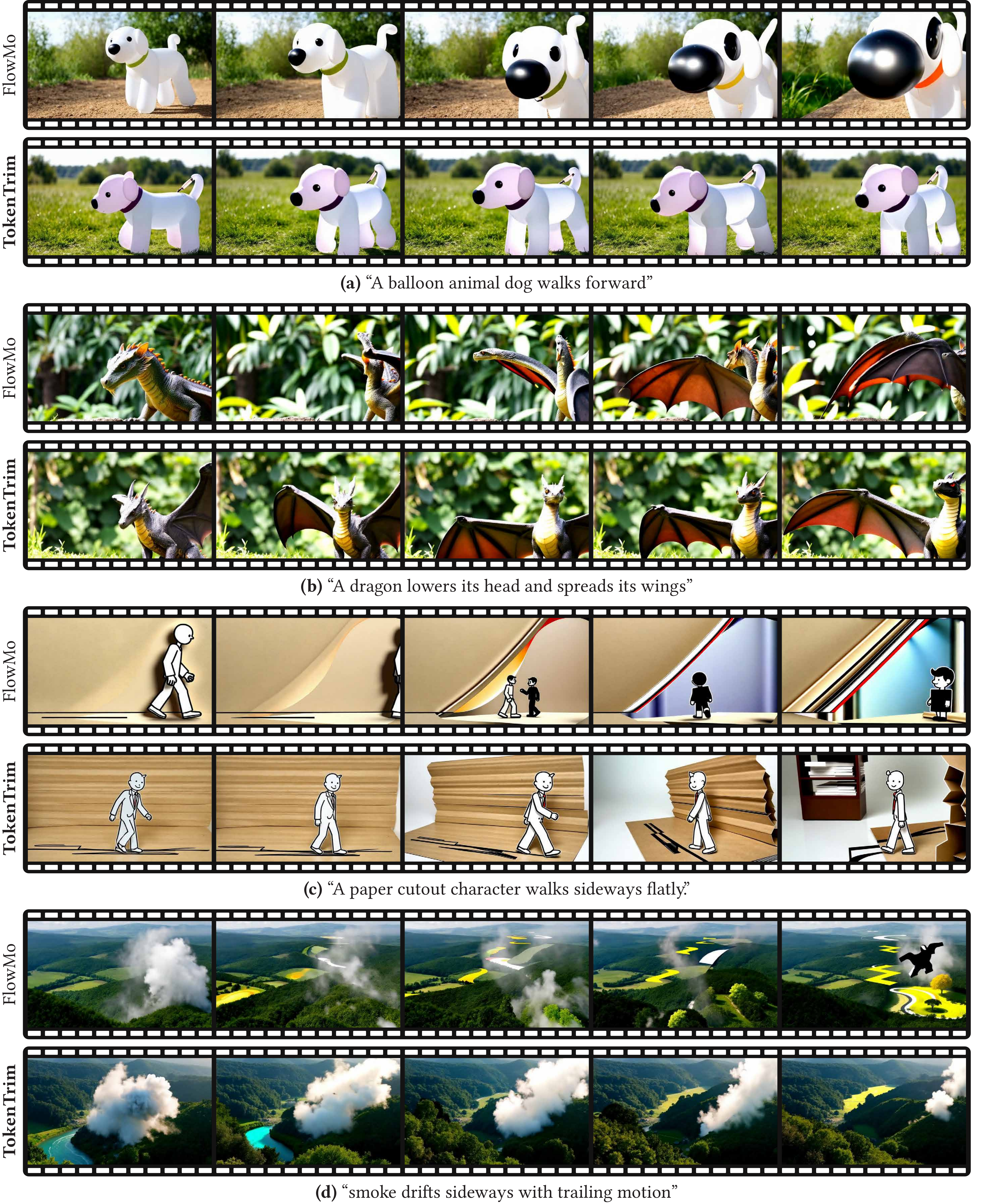}
  \caption{Text-to-video results from FlowMo \cite{shaulov2025flowmo} and TokenTrim.
  For additional qualitative results see App. ~\ref{app:additional_qualitative_flowmo}.}
  \label{fig:qualitative_flowmo_tokentrim}
  \vspace{-15pt}

\end{figure*}

\section{Method}

We propose an inference-time framework for stabilizing long-horizon auto-regressive video diffusion via latent-domain token pruning.
Our method is designed to operate on auto-regressive video diffusion models that condition future generation on previously generated frames via causal self-attention and a temporal key-value (KV) cache, such as Self Forcing~\cite{huang2025self} and Rolling Forcing~\cite{liu2025rolling}.
{The core mechanism of the proposed method focuses on identifying and removing unstable latent conditioning tokens during auto-regressive inference.} Algorithm~\ref{alg:tokentrim} outlines a single TokenTrim step.

\subsection{Motion-Stabilized Initialization}
\label{sec:motion_stabilized_initialization}

The quality of the first batch of generated frames is critical in auto-regressive video generation, as errors introduced at this stage propagate and amplify over time. To reduce early-stage corruption, we generate the first batch of frames using \emph{FlowMo}~\cite{shaulov2025flowmo}, which introduces motion-aware variance guidance at inference time. It encourages temporally coherent motion and reduces initial artifacts by guiding the denoising trajectory toward stable motion patterns. In our framework, FlowMo is applied only to the first batch of frames, producing a stable latent anchor. All subsequent batches are generated using the base auto-regressive model.

\subsection{Latent Summary Construction}
Let $\mathcal{X}_{t-1} = \{x_{t-1}^{(f)}\}_{f=1}^{F}$ denote the previously generated batch of $F$ frames at auto-regressive step $t-1$, where $x_{t-1}^{(f)} \in \mathbb{R}^{H \times W \times 3}$  denotes a single video frame. Each frame is encoded into a set of spatial latent tokens

\begin{equation}
\mathbf{Z}_{t-1}^{(f)} = E(x_{t-1}^{(f)}) \in \mathbb{R}^{N \times D},
\end{equation}
where $E(\cdot)$ denotes the encoder, $N$ is the number of spatial tokens, and $D$ is the latent dimension. We construct a latent \emph{summary frame} by averaging across the temporal dimension (frame index):
\begin{equation}
\mathbf{Z}_{t-1} = \frac{1}{F} \sum_{f=1}^{F} \mathbf{Z}_{t-1}^{(f)} \in \mathbb{R}^{N \times D}.
\end{equation}

After generating a candidate current batch $\mathcal{X}_{t}$, we treat it in the same manner as $\mathcal{X}_{t-1}$ and compute a corresponding latent summary $\mathbf{Z}_{t} \in \mathbb{R}^{N \times D}$. Averaging across frames ensures that $\mathbf{Z}_{t-1}$ and $\mathbf{Z}_{t}$ share identical spatial token structure, enabling direct token-wise comparison.

\begin{wrapfigure}{r}{0.52\textwidth}
\vspace{-2\baselineskip}
\begin{minipage}{\linewidth}
\begin{algorithm}[H]
\footnotesize
\caption{A Single TokenTrim Step}
\label{alg:tokentrim}
\textbf{Input:} Previous batch frames $\mathcal{X}_{t-1}$, candidate current batch frames $\mathcal{X}_{t}$, encoder $E$,
temporal KV cache $(\mathbf{K}_{\mathrm{cache}}, \mathbf{V}_{\mathrm{cache}})$, running drift statistics $(\mu_t, \sigma_t)$,
pruning fraction $p$, sensitivity $\lambda$, warm-up length $T_{\mathrm{warm}}$.\\
\textbf{Output:} Accepted current batch $\mathcal{X}_{t}^{\star}$ and updated KV cache.
\begin{algorithmic}[1]
\State $\mathbf{Z}_{t-1} \gets \frac{1}{F} \sum_{f=1}^{F} E(x_{t-1}^{(f)}) \in \mathbb{R}^{N \times D}$
\State $\mathbf{Z}_{t} \gets \frac{1}{F} \sum_{f=1}^{F} E(x_{t}^{(f)}) \in \mathbb{R}^{N \times D}$
\State $d_i \gets \|\mathbf{Z}_{t}(i) - \mathbf{Z}_{t-1}(i)\|_2 \quad \forall i \in \{1,\dots,N\}$
\State $\mathcal{S}_t \gets \text{TopIndices}(\{d_i\}_{i=1}^{N}, \lceil pN \rceil)$
\State $D_t \gets \frac{1}{|\mathcal{S}_t|}\sum_{i \in \mathcal{S}_t} d_i$
\If{$t \le T_{\mathrm{warm}}$}
    \State $\mathcal{X}_{t}^{\star} \gets \mathcal{X}_{t}$
    \State \textbf{UpdateStatsAndCache}$(\mathcal{X}_{t}^{\star}, \mathbf{K}_{\mathrm{cache}}, \mathbf{V}_{\mathrm{cache}})$
    \State \textbf{Return} $(\mathcal{X}_{t}^{\star}, \mathbf{K}_{\mathrm{cache}}, \mathbf{V}_{\mathrm{cache}})$
\EndIf
\If{$D_t \le \mu_t + \lambda \sigma_t$}
    \State $\mathcal{X}_{t}^{\star} \gets \mathcal{X}_{t}$
    \State \textbf{UpdateStatsAndCache}$(\mathcal{X}_{t}^{\star}, \mathbf{K}_{\mathrm{cache}}, \mathbf{V}_{\mathrm{cache}})$
    \State \textbf{Return} $(\mathcal{X}_{t}^{\star}, \mathbf{K}_{\mathrm{cache}}, \mathbf{V}_{\mathrm{cache}})$
\Else
    \State $\mathbf{m}_t(i) \gets \mathbb{1}[i \notin \mathcal{S}_t]$
    \State $\tilde{\mathbf{K}} \gets \mathbf{K}_{\mathrm{cache}}[\mathbf{m}_t],\;
           \tilde{\mathbf{V}} \gets \mathbf{V}_{\mathrm{cache}}[\mathbf{m}_t]$
    \State $\mathcal{X}_{t}^{\star} \gets \textsc{GenerateBatch}(\tilde{\mathbf{K}}, \tilde{\mathbf{V}})$
    \State \textbf{UpdateStatsAndCache}$(\mathcal{X}_{t}^{\star}, \tilde{\mathbf{K}}, \tilde{\mathbf{V}})$
    \State \textbf{Return} $(\mathcal{X}_{t}^{\star}, \tilde{\mathbf{K}}, \tilde{\mathbf{V}})$
\EndIf
\end{algorithmic}
\end{algorithm}
\end{minipage}
\vspace{-2\baselineskip}
\end{wrapfigure}

\subsection{Per-Token Latent Drift Estimation}

We estimate instability in the auto-regressive context by computing a per-token drift score using latent-space subtraction. For each spatial token index $i \in \{1, \dots, N\}$, the drift score is defined as
\begin{equation}
d_i = \left\| \mathbf{Z}_{t}(i) - \mathbf{Z}_{t-1}(i) \right\|_2.
\end{equation}
This measure captures deviations in semantic and structural features at the patch level and serves as an indicator of latent instability between consecutive auto-regressive steps.

This comparison is performed between latent summaries that share the same tokenization grid and are computed at the same spatial resolution; moderate camera motion is therefore reflected as smooth, coherent changes across neighboring tokens rather than isolated high-magnitude drift, while corrupted or unstable regions induce localized spikes in $d_i$.

\subsection{Drift Severity and Trigger Criterion}

Rather than pruning unconditionally, we first assess whether the current batch exhibits abnormal drift. Let $p \in (0,1)$ denote the pruning fraction, and let $N$ be the number of spatial latent tokens. For each auto-regressive step $t$, we rank the per-token drift values $\{d_i\}_{i=1}^{N}$ and define
\[
\mathcal{S}_t \subseteq \{1,\dots,N\}, \qquad |\mathcal{S}_t| = \lceil p\cdot N \rceil,
\]
as the set of indices corresponding to the top $p \cdot N$ spatial tokens with the largest drift scores.

We define a scalar \emph{drift severity score} $D_t \in \mathbb{R}$ as the mean drift over these tokens:
\begin{equation}
D_t = \frac{1}{|\mathcal{S}_t|} \sum_{i \in \mathcal{S}_t} d_i,
\qquad d_i \in \mathbb{R}_{\ge 0}.
\end{equation}

To obtain an adaptive threshold, we maintain running statistics over previously accepted batches. A batch $\mathcal{X}_\tau$ is considered accepted if it is finalized (completed a tokentrim step, see Alg.~\ref{alg:tokentrim}) and appended to the auto-regressive context.
Let
\[
\mathcal{A}_t = \{\tau < t \mid \mathcal{X}_\tau \text{ was accepted}\}
\]
denote the index set of such accepted auto-regressive steps prior to $t$, and let $|\mathcal{A}_t|$ be its cardinality. We define the running mean $\mu_t \in \mathbb{R}$ and standard deviation $\sigma_t \in \mathbb{R}_{\ge 0}$ of the drift severity as
\begin{equation}
\mu_t = \frac{1}{|\mathcal{A}_t|} \sum_{\tau \in \mathcal{A}_t} D_\tau,
\qquad
\sigma_t =
\sqrt{
\frac{1}{|\mathcal{A}_t|}
\sum_{\tau \in \mathcal{A}_t}
\left(D_\tau - \mu_t\right)^2
}.
\end{equation}

We trigger a pruning intervention when the current drift severity exceeds the adaptive threshold:
\begin{equation}
D_t > \mu_t + \lambda \sigma_t,
\end{equation}
where $\lambda > 0$ is a sensitivity hyperparameter controlling the strictness of the trigger (we use $\lambda = 2.0$).

During the first $T_{\mathrm{warm}}$ auto-regressive steps, the statistics $\mu_t$ and $\sigma_t$ may be unreliable due to limited history. We therefore disable pruning during this warm-up phase and only accumulate drift statistics.

\subsection{Hard Pruning and Regeneration}

When the drift criterion is exceeded, we regenerate the current batch with \emph{hard pruning} applied to the auto-regressive context. Specifically, we remove from the temporal KV cache all token positions whose spatial indices belong to $\mathcal{S}_t$. Let $\mathbf{m}_t \in \{0,1\}^{N}$ denote the spatial pruning mask:
\begin{equation}
\mathbf{m}_t(i) =
\begin{cases}
0, & i \in \mathcal{S}_t, \\
1, & \text{otherwise}.
\end{cases}
\end{equation}
We apply this mask to the cached keys and values (conceptually along the token dimension):
\begin{equation}
\tilde{\mathbf{K}} = \mathbf{K}_{\text{cache}}[\mathbf{m}_t], 
\qquad
\tilde{\mathbf{V}} = \mathbf{V}_{\text{cache}}[\mathbf{m}_t],
\end{equation}
where $\mathbf{K}_{\text{cache}}$ and $\mathbf{V}_{\text{cache}}$ denote the cached keys and values corresponding to previously generated frames. The pruned tokens $(\tilde{\mathbf{K}}, \tilde{\mathbf{V}})$ are used to condition a second generation attempt of the current batch. 
If the drift criterion is not exceeded, the generated batch is accepted without pruning, see Fig ~\ref{fig:method_figure}.
{Formally, the current batch $\mathcal{X}_t$ is accepted as the final output for step $t$ and appended to the auto-regressive context without any pruning or regeneration.}
For stability and efficiency, we limit the procedure to at most $R$ regeneration attempts per batch (we use $R=1$).
That is, when pruning is triggered, the batch is regenerated at most once using the pruned KV cache; if the regenerated batch still violates the drift criterion, it is accepted as-is to avoid unbounded regeneration loops.

\subsection{Integration with Self Forcing and Rolling Forcing}

In \emph{Self Forcing}~\cite{huang2025self}, which employs a rolling KV cache during inference, latent drift estimation and hard pruning are applied before appending new KV entries to the cache. In \emph{Rolling Forcing}~\cite{liu2025rolling}, which separates a global anchor cache (derived from initial frames) from a recent temporal context cache, hard pruning is applied exclusively to the recent context tokens.

\begin{figure*}[t]
  \centering
  \includegraphics[width=\textwidth]{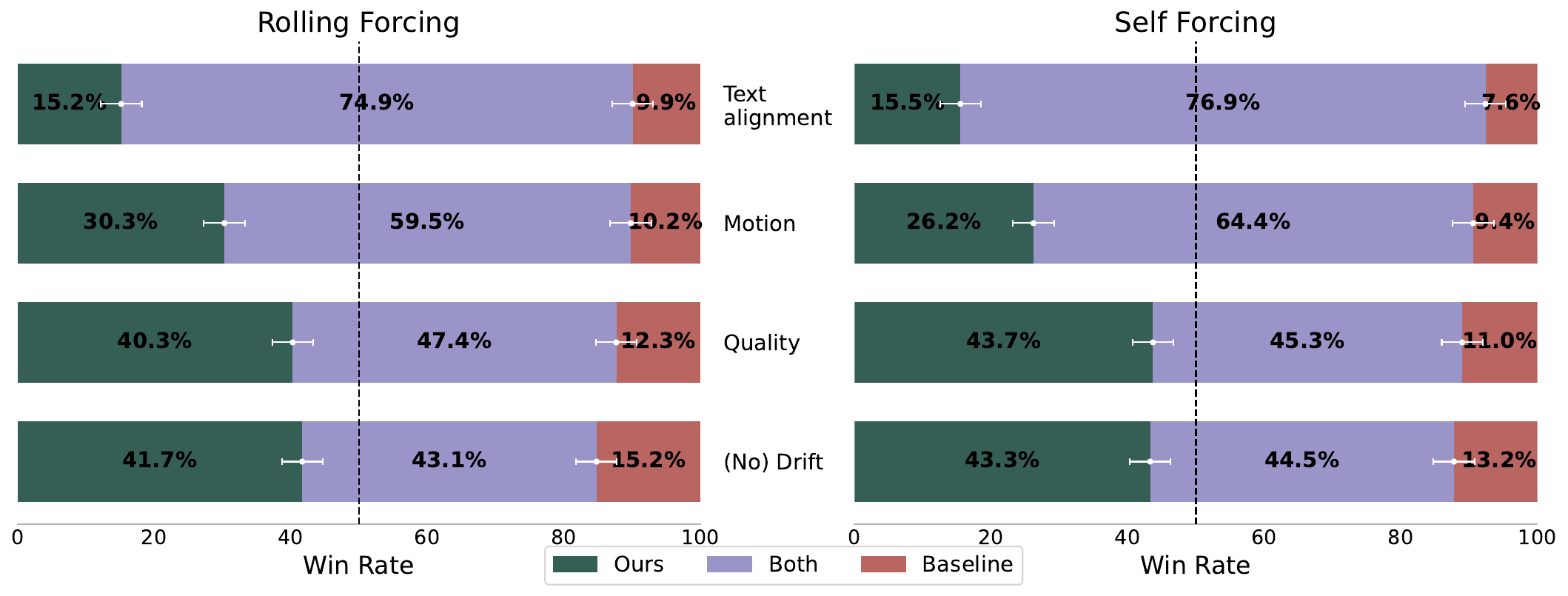}
  \caption{
    Human preference study conducted on Rolling Forcing~\cite{liu2025rolling} (left) and Self Forcing~\cite{huang2025self} (right) using VideoJAM-bench~\cite{chefer2025videojam}. TokenTrim is consistently preferred in terms of drift reduction, motion quality, and overall visual quality, while preserving text–video alignment. Error bars indicate 95\% confidence intervals computed via Dirichlet sampling with Laplace smoothing.
  }
  \label{fig:user_study}
\end{figure*}

Overall, our method addresses temporal drift by operating entirely in the latent space; it requires no additional training or supervision, and it introduces almost no computational overhead.

\section{Experiments}
We present a comprehensive evaluation of TokenTrim in the setting of long-horizon auto-regressive text-to-video generation. Our experiments examine whether selectively pruning unstable latent tokens at inference time can mitigate error accumulation across generation steps, while preserving visual quality, motion realism, and semantic alignment. We evaluate TokenTrim both quantitatively, using VBench metrics, and qualitatively, via visual comparisons and human preference studies. All evaluations are conducted under the same experimental conditions and include comparisons against both baseline auto-regressive methods and inference-time methods.

\paragraph{\textbf{Implementation details.}\quad}
We evaluate TokenTrim on Rolling Forcing\cite{liu2025rolling} and Self Forcing\cite{huang2025self}, two auto-regressive inference strategies built on the Wan2.1-1.3B text-to-video model~\cite{wan2025wan}. TokenTrim operates purely at inference time and introduces no changes to the underlying model weights or denoising dynamics.

TokenTrim monitors latent drift between consecutive generated batches and triggers hard pruning when abnormal drift is detected. Unless otherwise stated, we use a pruning ratio of p=0.1, a drift threshold parameter $\lambda=2.0$, and a warm-up period of $T_{\mathrm{warm}}=2$ batches. These parameters are specific to TokenTrim and are independent of the underlying generation model or inference strategy. When pruning is triggered, the current batch is regenerated once using the pruned temporal KV cache.

In the FlowMo-adapted setup, FlowMo\cite{shaulov2025flowmo} is applied during the generation of all the batches using its default inference-time settings. In contrast, when combined with TokenTrim, FlowMo is applied only to the first batch to produce a stable initialization, as described in Sec.~\ref{sec:motion_stabilized_initialization}, and is disabled for subsequent batches.

All experiments are conducted on a single NVIDIA H100 GPU. Rolling Forcing and Self Forcing generate 30-second videos at 16 FPS and a resolution of $832 \times 480$.


\subsection{Quantitative Results}
\label{token_trim_quantitative_results}

We evaluate TokenTrim using the VBench benchmark~\cite{huang2024vbench}, which provides a comprehensive suite of motion, quality, and semantic metrics for text-to-video generation. We report both per-dimension scores and aggregated metrics, including Semantic Score, Quality Score, and the overall Final Score. Comparisons are made against the baseline auto-regressive methods, Rolling Forcing~\cite{liu2025rolling} and Self Forcing\cite{huang2025self} as well as FlowMo~\cite{shaulov2025flowmo}, using identical prompts and generation settings.

\begin{wraptable}{r}{0.52\textwidth}
\footnotesize
\renewcommand{\arraystretch}{0.85}
\setlength{\tabcolsep}{4pt}
\caption{\textbf{VBench evaluation results.}
Comparison of baseline auto-regressive inference, FlowMo, and TokenTrim across aggregated VBench scores.}
\label{tab:vbench_tokentrim_flowmo}
\begin{tabular}{lccc}
\toprule
\textbf{Model} & \textbf{Semantic} & \textbf{Quality} & \textbf{Final} \\
\midrule
Rolling Forcing & 68.52\% & 81.72\% & 75.12\% \\
+ FlowMo        & 69.53\% & 82.09\% & 75.81\% \\
+ TokenTrim     & \textbf{72.05\%} & \textbf{87.30\%} & \textbf{79.67\% \highlight{(+4.55\%)}} \\
\midrule
Self Forcing    & 68.98\% & 82.89\% & 75.93\% \\
+ FlowMo        & 68.25\% & 83.85\% & 76.05\% \\
+ TokenTrim     & \textbf{73.89\%} & \textbf{89.79\%} & \textbf{81.84\% \highlight{(+5.91\%)}} \\
\bottomrule
\end{tabular}

\end{wraptable}

\paragraph{\textbf{Automatic metrics.}\quad}
Tab.~\ref{tab:vbench_tokentrim_flowmo} reports aggregated VBench results comparing TokenTrim against the baseline auto-regressive methods (Rolling Forcing and Self Forcing) as well as the inference-time method FlowMo, under identical generation settings. We enclose the aggregated metrics, which constitute an average of all the benchmark dimensions, and measure the overall quality of the generations. A full breakdown of all metrics is provided in App~\ref{app:VBench_Metrics_Breakdown}.

When applied on top of Rolling Forcing, TokenTrim yields substantial improvements across all aggregated VBench metrics. The \textbf{Final Score} increases from 75.12\% to 79.67\% (+4.55\%), supported by gains in \textbf{Quality Score} (+5.58\%) and \textbf{Semantic Score} (+3.53\%). These improvements indicate that selectively pruning unstable latent tokens significantly enhances both perceptual quality and semantic consistency over long auto-regressive rollouts.

A similar trend is observed under Self Forcing. TokenTrim improves the \textbf{Final Score} from 75.93\% to 81.84\% (+5.91\%), accompanied by strong gains in \textbf{Quality Score} (+6.90\%) and \textbf{Semantic Score} (+4.91\%). Notably, the magnitude of these improvements is even larger than under Rolling Forcing, highlighting TokenTrim’s effectiveness across different auto-regressive inference strategies.

TokenTrim also consistently outperforms FlowMo at the aggregate level. Under Self Forcing, TokenTrim raises the \textbf{Final Score} to 81.84\% (+5.91\%), whereas FlowMo yields only a marginal increase to 76.05\% (+0.12\%). This gap is further reflected in the \textbf{Semantic Score}, which improves substantially with TokenTrim (+4.91\%) but slightly degrades with FlowMo (-0.73\%), as well as in the \textbf{Quality Score}, where TokenTrim achieves a +6.90\% gain compared to FlowMo’s +0.96\%.

Overall, these aggregated results demonstrate that TokenTrim provides a significantly stronger and more reliable improvement in semantic fidelity and visual quality than both baseline auto-regressive methods and inference-time motion guidance, leading to higher overall generation quality in long-horizon video synthesis.

\paragraph{\textbf{Inference-Time Overhead.}\quad}
Averaged over 128 generated 30s videos, TokenTrim increases wall-clock runtime by $\times1.08$ relative to the Rolling Forcing baseline.
In contrast, applying the FlowMo-adapted setup incurs a substantially larger cost, resulting in a $\times2.18$ slowdown over Rolling Forcing.

\subsection{Qualitative Results}
We qualitatively compare TokenTrim against the baseline auto-regressive methods and FlowMo using long-horizon generations exceeding one minute. Representative examples are shown in Fig. ~\ref{fig:teaser} and Fig. ~\ref{fig:qualitative}.

Across all examples, TokenTrim maintains stable object identities, colors, and structure over time, whereas the baseline methods and FlowMo exhibit progressive degradation. Common failure modes include color shifts, structural distortions, background corruption, and identity drift. For example, in Fig. ~\ref{fig:qualitative}(a), the baseline Rolling Forcing model produces a Pikachu character with missing or duplicated limbs over time. In Fig. ~\ref{fig:teaser}(a,b), baseline generations show noticeable color drift and texture degradation, while TokenTrim preserves consistent appearance. Similarly, artifacts such as lens flare accumulation and background warping are visible in Fig. ~\ref{fig:qualitative}(b–d) for the baseline and FlowMo, but are largely absent when using TokenTrim. Additional qualitative comparisons between TokenTrim and Rolling Forcing are provided in App.~\ref{app:additional_qualitative_rolling}, with corresponding results for Self Forcing reported in App.~\ref{app:additional_qualitative_self}.

\begin{wrapfigure}[20]{r}{0.48\columnwidth} 
  \centering
  \includegraphics[width=\linewidth]{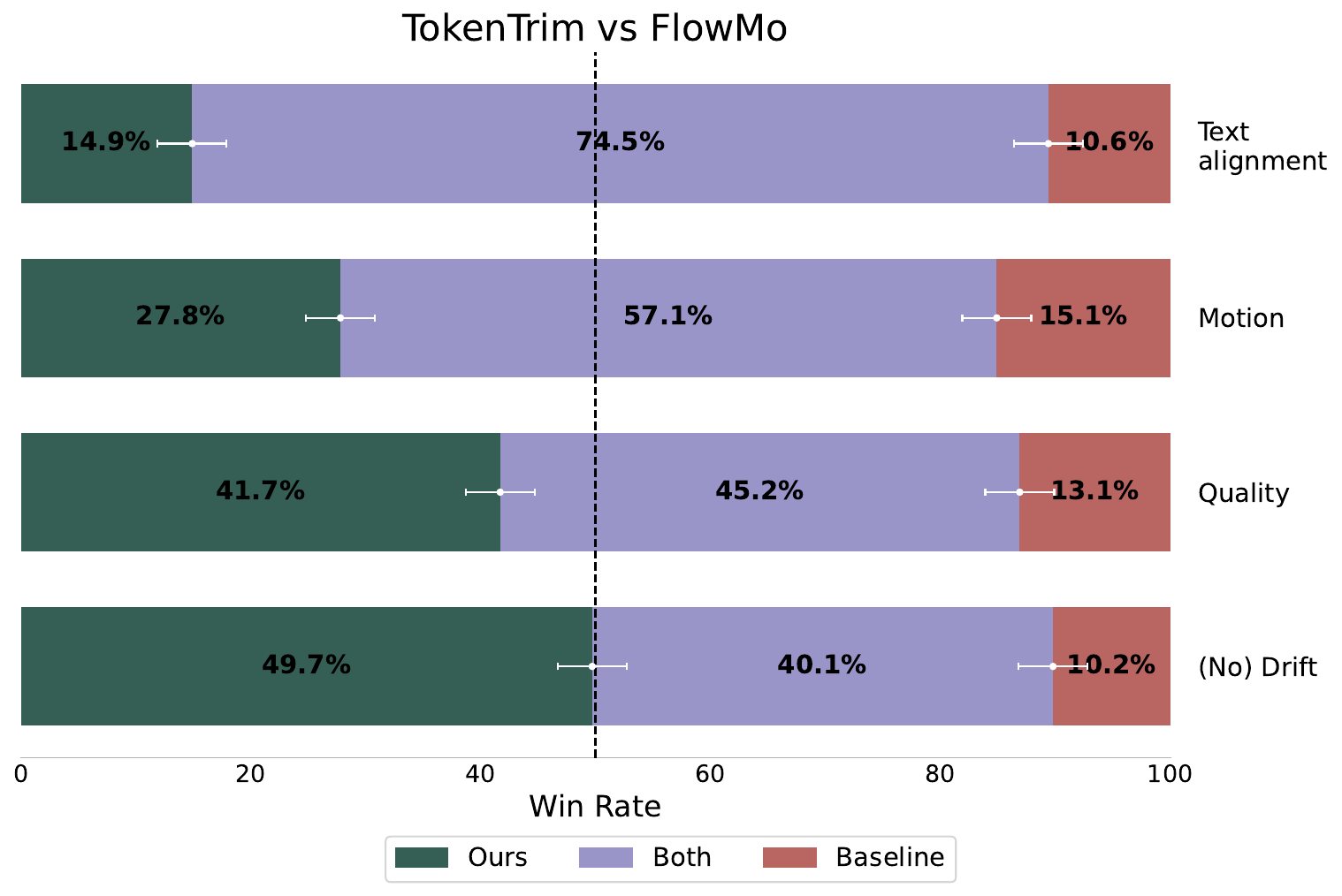}
  \caption{\footnotesize Human preference study on VideoJAM-Bench~\cite{chefer2025videojam}. Win rates comparing TokenTrim against FlowMo~\cite{shaulov2025flowmo} under Rolling Forcing. Error bars: 95\% CIs via Dirichlet sampling with Laplace smoothing.}
  \label{fig:user_study_flowmo}
\end{wrapfigure}

Fig. ~\ref{fig:qualitative_flowmo_tokentrim} directly compares TokenTrim with FlowMo under identical settings. While FlowMo often produces plausible short-term motion, it exhibits gradual structural drift over long horizons. For example, in Fig. ~\ref{fig:qualitative_flowmo_tokentrim}(a), the balloon animal dog generated with FlowMo progressively deforms, most noticeably in the snout and head shape, whereas TokenTrim preserves consistent geometry and proportions. In Fig. ~\ref{fig:qualitative_flowmo_tokentrim}(b), FlowMo introduces subtle distortions in the dragon’s wings and head during the rollout, while TokenTrim maintains coherent anatomy and smooth articulation. Similar effects are observed in Fig. ~\ref{fig:qualitative_flowmo_tokentrim}(c,d), where FlowMo suffers from background warping and implied motion drift, whereas TokenTrim maintains stable silhouettes and coherent global motion. Additional qualitative comparisons between TokenTrim and FlowMo are provided in App.~\ref{app:additional_qualitative_flowmo}.

These examples illustrate that TokenTrim stabilizes long-horizon generation by preventing corrupted latent tokens from repeatedly influencing future steps, resulting in videos that remain visually coherent throughout extended rollouts. 

\subsection{User Study}
We conduct a human preference study using prompts from the VideoJAM benchmark ~\cite{chefer2025videojam}. For each prompt, participants are shown paired videos generated under identical settings, differing only in the inference-time method (baseline, FlowMo, or TokenTrim). Video order is randomized to avoid positional bias. Each pair is evaluated by five independent annotators, resulting in 640 responses per baseline. Participants evaluate videos along four criteria: text–video alignment, aesthetic quality, motion coherence, and temporal drift  (see App. ~\ref{app:user_study}).

Figure~\ref{fig:user_study} reports human preference results comparing TokenTrim against the baseline under both Rolling Forcing (left) and Self Forcing (right). Under Rolling Forcing, TokenTrim achieves a preference rate of 15.2\% in text–video alignment, compared to 9.9\% for the baseline, indicating that pruning unstable tokens does not harm semantic consistency. For motion coherence, TokenTrim is preferred in 30.3\% of cases, nearly three times higher than the baseline (10.2\%), demonstrating a clear improvement in temporally plausible motion. A similar trend is observed for aesthetic quality, where TokenTrim reaches 40.3\% preference versus 12.3\% for the baseline. The largest improvement is observed in temporal drift, with TokenTrim attaining 41.7\% preference compared to 15.2\% for the baseline, highlighting its effectiveness in maintaining stable visual characteristics over long rollouts.
A consistent pattern emerges under Self Forcing. TokenTrim improves text–video alignment to 15.5\%, compared to 7.6\% for the baseline. For motion coherence, TokenTrim achieves a preference rate of 26.2\%, substantially higher than the baseline (9.4\%). In aesthetic quality, TokenTrim is preferred in 43.7\% of comparisons, compared to 11.0\% for the baseline. Finally, for temporal drift, TokenTrim reaches 43.3\% preference, significantly outperforming the baseline (13.2\%), confirming that the benefits of TokenTrim generalize across auto-regressive inference strategies.

Fig. ~\ref{fig:user_study_flowmo} reports a direct human preference comparison between TokenTrim and FlowMo. Across all evaluated criteria, TokenTrim consistently receives more favorable votes than FlowMo.
For text–video alignment, TokenTrim is preferred in 14.9\% of comparisons, compared to 10.6\% for FlowMo, indicating that TokenTrim provides stronger semantic alignment between the generated videos and the input prompts. In terms of motion coherence, TokenTrim achieves a higher preference rate of 27.8\%, surpassing FlowMo’s 15.1\%, demonstrating that TokenTrim exceeds FlowMo’s motion quality despite not explicitly optimizing motion dynamics.
TokenTrim also shows a substantial advantage in aesthetic quality, receiving 41.7\% of preferences compared to 13.1\% for FlowMo, reflecting consistently stronger visual appearance. The largest gap is observed in temporal drift, where TokenTrim attains a preference rate of 49.7\%, dramatically outperforming FlowMo’s 10.2\%. This result highlights TokenTrim’s superior ability to preserve stable structure, appearance, and identity over long generation horizons.
Overall, the user study confirms that TokenTrim produces videos that are perceived as more stable, coherent, and visually consistent over long horizons.

\begin{table}[h]
\centering
\renewcommand{\arraystretch}{0.9}
\caption{\textbf{TokenTrim Ablation on Rolling Forcing.}
Ablation study of TokenTrim (TT) under different pruning ratios and initialization settings.
The full TokenTrim method uses 10\% pruning with FlowMo initialization.
Columns correspond to the aggregated Semantic Score (Semantic), Quality Score (Quality), and the overall Final Score (Final) as defined by VBench.}
\label{tab:vbench_tokentrim_ablation}
\begin{tabular}{lccc}
\toprule
\textbf{Model} &
\textbf{Semantic} &
\textbf{Quality} &
\textbf{Final} \\
\midrule
\textbf{Full TT Method} &
\textbf{72.05\%} &
\textbf{87.30\%} &
\textbf{79.67\%} \\
\midrule
TT 5\% Pruning &
70.78\% &
85.92\% &
78.35\% \textcolor{red}{(-1.32\%)} \\

TT 20\% Pruning &
64.22\% &
72.29\% &
68.25\% \textcolor{red}{(-11.87\%)} \\

TT w/o FlowMo Init &
71.17\% &
83.49\% &
77.33\% \textcolor{red}{(-2.34\%)} \\
\bottomrule
\end{tabular}
\end{table}

\subsection{Ablation Study}

We conduct an ablation study to analyze the contribution of each component in TokenTrim and to evaluate the effect of different pruning strategies. Tab.~\ref{tab:vbench_tokentrim_ablation} reports aggregated VBench results for a set of TokenTrim variants evaluated on top of Rolling Forcing. In particular, we analyze alternative pruning strategies, including a variant that disables the FlowMo-based motion-stabilized initialization, and examine sensitivity to the pruning ratio by testing fixed pruning rates of $5\%$ and $20\%$, instead of the $10\%$ we use throughout all other experiments.
Pruning $5\%$ of the most unstable tokens results in a moderate degradation over the baseline TokenTrim method, achieving a final score of $78.35\%$ ($-1.32\%$ compared to the baseline TokenTrim method), while aggressive pruning at $20\%$ severely degrades performance, reducing the final score to $68.25\%$ ($-11.87\%$). These results highlight that excessive removal of contextual tokens disrupts semantic continuity and visual quality. In another ablation, the FlowMo is removed from the initialization of the first batch (FlowMo is not used in the subsequent batches). This leads to a final score of $77.33\%$ which is $-2.34\%$ less than the full TokenTrim method, demonstrating that motion-stabilized initialization provides a complementary benefit by improving early coherence. Interestingly, the degradation that occures in this ablation, is greater than the benefit of FlowMo to the baseline Rolling Forcing method (see Tab.~\ref{tab:vbench_tokentrim_flowmo}), demonstrating that TokenTrim leads to better utilization of FlowMo than the baseline Rolling Forcing. A full breakdown of all metrics is provided in App.~\ref{app:ablation_study}.

\section{Limitations \& Future Work}

TokenTrim operates purely at inference time and leaves model parameters unchanged; therefore, its gains are ultimately bounded by the capabilities and biases of the underlying video diffusion backbone and its training data. When the base model persistently struggles to represent an object, preserve identity, or produce stable motion, TokenTrim can primarily attenuate error propagation rather than fully correct the generation. Moreover, our current implementation uses a fixed hard-pruning budget (e.g., pruning a constant fraction or top-$k$ tokens per step). Although this design is lightweight and adds negligible overhead, a single global pruning setting may be suboptimal across prompts, content types, and rollout lengths. An important direction for future work is to make pruning adaptive, i.e., dynamically choosing the pruning rate and structure at each step using drift statistics or uncertainty estimates so that difficult sequences receive stronger suppression while stable sequences retain richer context, especially over long rollouts.

\section{Conclusions}

Can we sustain long-video generation by simply knowing what to forget? In this work, we address the pervasive issue of temporal drift not by adding more capacity or retraining, but by pruning tokens at inference time. Since the uncontrolled accumulation of corrupted tokens in the auto-regressive context is a primary driver of semantic collapse, TokenTrim computes a drift score for the model's latent representation and distinguishes between stable context and hallucinated artifacts. We demonstrate that pruning these unstable tokens allows the model to maintain its connection to reliable anchors, effectively breaking the feedback loop of error accumulation. This approach transforms the KV-cache from a passive history log into an active self-correcting memory mechanism. Our results suggest that future progress in infinite video synthesis lies not only in how much context a model can remember, but also in its ability to selectively discard the noise that threatens to distort reality.

\section{Acknowledgments}
This work was supported by a grant from the Tel Aviv University Center for AI and Data Science (TAD).

\bibliographystyle{unsrt}  


\newpage

\appendix

\section{VBench Metrics Breakdown}
\label{app:VBench_Metrics_Breakdown}

We evaluate the impact of \textbf{TokenTrim} on both Rolling Forcing \cite{liu2025rolling} and Self Forcing \cite{huang2025self} using the VBench benchmark, with full per-dimension results reported in Tab.~\ref{tab:full_vbench_unified}.

The VBench metrics~\cite{huang2024vbench} consistently demonstrate that TokenTrim yields substantial improvements across a wide range of dimensions for both auto-regressive baselines. Notably, TokenTrim improves performance in nearly all motion- and stability-related metrics, indicating effective mitigation of error accumulation across auto-regressive steps rather than localized, within-chunk refinement.

For \textbf{Rolling Forcing}, TokenTrim provides clear gains in key motion coherence metrics, improving \textbf{Temporal Flickering} by +2.12\% (from 96.77\% to 98.89\%), \textbf{Motion Smoothness} by +1.81\%, and \textbf{Dynamic Degree} by +2.54\%. These improvements indicate that TokenTrim enhances temporal stability while preserving motion complexity, rather than suppressing dynamics. TokenTrim further improves higher-level consistency metrics such as \textbf{Human Action} (+4.19\%), \textbf{Scene} (+2.98\%), and \textbf{Overall Consistency} (+3.45\%), reflecting stronger long-horizon semantic coherence.

A similar pattern is observed for \textbf{Self Forcing}. TokenTrim improves \textbf{Temporal Flickering} (+0.03\%), \textbf{Motion Smoothness} (+1.01\%), and \textbf{Dynamic Degree} (+1.22\%), while also yielding consistent gains in \textbf{Human Action}, \textbf{Appearance Style} (+2.07\%), and \textbf{Overall Consistency} (+0.64\%). Importantly, these improvements arise despite Self Forcing already employing stronger temporal anchoring, highlighting that TokenTrim addresses a complementary failure mode—namely, the repeated reuse of corrupted latent tokens across auto-regressive steps.

In the aggregated metrics, \emph{TokenTrim delivers strong and consistent improvements for both baselines}. For Rolling Forcing, TokenTrim increases the \textbf{Semantic Score} by +3.53\% and the \textbf{Quality Score} by +5.58\%, resulting in a \textbf{Final Score} improvement of +4.55\%. For Self Forcing, the gains are even larger, with +4.91\% in Semantic Score, +6.90\% in Quality Score, and a \textbf{Final Score} increase of +5.91\%. These consistent improvements across both architectures confirm that TokenTrim robustly mitigates long-horizon drift by controlling context reuse, leading to higher-quality and more stable video generation over extended rollouts.

\renewcommand{\arraystretch}{0.9}
\begin{table*}[t!]
\centering
\footnotesize
\caption{\textbf{VBench evaluation across all dimensions.}
Comparison of Rolling Forcing and Self Forcing with TokenTrim and FlowMo.}
\label{tab:full_vbench_unified}
\vspace{-4pt}
\setlength{\tabcolsep}{4pt}

\begin{tabular*}{\textwidth}{@{\extracolsep{\fill}} l c c c | c c c @{}}
\toprule
\textbf{Dimension} &
\textbf{Rolling Forcing} &
\textbf{+ TokenTrim} &
\textbf{+ FlowMo} &
\textbf{Self Forcing} &
\textbf{+ TokenTrim} &
\textbf{+ FlowMo} \\
\midrule
Subject Consistency        & 91.92\% & 90.12\% & \textbf{91.94\%} & \textbf{88.29\%} & 85.54\% & 87.99\% \\
Background Consistency     & 92.87\% & \textbf{93.54\%} & 91.27\% & 89.53\% & \textbf{89.72\%} & 87.12\% \\
Temporal Flickering        & 96.77\% & \textbf{98.89\%} & 95.21\% & 98.90\% & \textbf{98.93\%} & 98.91\% \\
Motion Smoothness          & 97.21\% & \textbf{99.02\%} & 98.03\% & 97.63\% & \textbf{98.64\%} & 97.95\% \\
Dynamic Degree             & 59.57\% & \textbf{62.11\%} & 56.07\% & 67.75\% & \textbf{68.97\%} & 65.46\% \\
Aesthetic Quality          & \textbf{64.19\%} & 63.95\% & 63.05\% & 61.06\% & \textbf{63.13\%} & 62.15\% \\
Imaging Quality            & 71.70\% & \textbf{71.91\%} & 68.62\% & 68.92\% & \textbf{69.17\%} & 66.99\% \\
Object Class               & 88.97\% & \textbf{89.63\%} & 89.33\% & 81.24\% & 80.30\% & \textbf{82.24\%} \\
Multiple Objects           & \textbf{68.31\%} & 65.29\% & 67.28\% & \textbf{63.98\%} & 60.37\% & 61.03\% \\
Human Action               & 75.81\% & \textbf{80.00\%} & 74.89\% & 83.72\% & \textbf{83.93\%} & 82.09\% \\
Color                      & \textbf{87.98\%} & 85.53\% & 86.20\% & 81.21\% & \textbf{81.28\%} & 79.83\% \\
Spatial Relationship       & 79.26\% & 75.09\% & \textbf{80.53\%} & 74.76\% & 76.02\% & \textbf{76.59\%} \\
Scene                      & 29.22\% & \textbf{32.20\%} & 30.85\% & \textbf{32.59\%} & 30.94\% & 29.25\% \\
Appearance Style           & \textbf{21.72\%} & 21.40\% & 19.56\% & 23.49\% & \textbf{25.56\%} & 24.15\% \\
Temporal Style             & 23.26\% & \textbf{25.49\%} & 25.12\% & \textbf{20.55\%} & 20.39\% & 20.02\% \\
Overall Consistency        & 24.92\% & \textbf{28.37\%} & 26.98\% & 25.12\% & \textbf{25.76\%} & 25.10\% \\
\midrule
Semantic Score             & 68.52\% & \textbf{72.05\%} & 69.53\% & 68.98\% & \textbf{73.89\%} & 68.25\% \\
Quality Score              & 81.72\% & \textbf{87.30\%} & 82.09\% & 82.89\% & \textbf{89.79\%} & 83.85\% \\
Final Score                & 75.12\% & \textbf{79.67\% \highlight{(+4.55\%)}} & 75.81\% & 75.93\% & \textbf{81.84\% \highlight{(+5.91\%)}} & 76.05\% \\
\bottomrule
\end{tabular*}

\vspace{-6pt}
\end{table*}

\renewcommand{\arraystretch}{0.9}
\begin{table*}[t!]
\centering
\footnotesize
\caption{VBench ablation across all dimensions.}
\label{tab:full_ablation}
\vspace{-4pt}
\setlength{\tabcolsep}{4pt}

\begin{tabular*}{\textwidth}{@{\extracolsep{\fill}} l c c c c c @{}}
\toprule
\textbf{Dimension} &
\textbf{Rolling Forcing} &
\textbf{+ Full TT Method} &
\textbf{+ TT 5\% Pruning} &
\textbf{+ TT 20\% Pruning} &
\textbf{+ TT w/o FlowMo init} \\
\midrule
Subject Consistency        & \textbf{91.92\%} & 90.12\% & 90.59\% & 85.31\% & 90.10\% \\
Background Consistency     & 92.87\% & \textbf{93.54\%} & 93.07\% & 78.02\% & 93.28\% \\
Temporal Flickering        & 96.77\% & \textbf{98.89\%} & 97.01\% & 89.83\% & 97.25\% \\
Motion Smoothness          & 97.21\% & \textbf{99.02\%} & 97.65\% & 85.52\% & 97.56\% \\
Dynamic Degree             & 59.57\% & \textbf{62.11\%} & 61.48\% & 55.97\% & 60.11\% \\
Aesthetic Quality          & 64.19\% & 63.95\% & \textbf{64.32\%} & 59.82\% & 62.54\% \\
Imaging Quality            & 71.70\% & 71.91\% & \textbf{72.81\%} & 62.68\% & 70.62\% \\
Object Class               & 88.97\% & \textbf{89.63\%} & 87.66\% & 72.11\% & 88.03\% \\
Multiple Objects           & \textbf{68.31\%} & 65.29\% & 64.13\% & 62.83\% & 64.42\% \\
Human Action               & 75.81\% & \textbf{80.00\%} & 77.29\% & 69.18\% & 78.95\% \\
Color                      & \textbf{87.98\%} & 85.53\% & 87.51\% & 64.32\% & 86.06\% \\
Spatial Relationship       & \textbf{79.26\%} & 75.09\% & 76.42\% & 68.04\% & 77.36\% \\
Scene                      & 29.22\% & \textbf{32.20\%} & 30.63\% & 29.75\% & 28.74\% \\
Appearance Style           & 21.72\% & 21.40\% & \textbf{21.98\%} & 11.86\% & 20.91\% \\
Temporal Style             & 23.26\% & \textbf{25.49\%} & 22.78\% & 15.29\% & 25.41\% \\
Overall Consistency        & 24.92\% & \textbf{28.37\%} & 25.65\% & 19.21\% & 26.31\% \\
\midrule
Semantic Score             & 68.52\% & \textbf{72.05\%} & 70.78\% & 59.22\% & 71.17\% \\
Quality Score              & 81.72\% & \textbf{87.30\%} & 85.92\% & 67.29\% & 83.49\% \\
Final Score                & 75.12\% & \textbf{79.67\% \highlight{(+4.55\%)}} &
                              78.35\% \highlight{(+3.23\%)} &
                              63.25\% \textcolor{red}{(-11.87\%)} &
                              77.33\% \highlight{(+2.21\%)} \\
\bottomrule
\end{tabular*}

\vspace{-6pt}
\end{table*}

We also compare \textbf{TokenTrim} against \textbf{FlowMo} using the VBench benchmark, with detailed per-dimension results reported in Tab.~\ref{tab:full_vbench_unified}.

The VBench metrics~\cite{huang2024vbench} clearly demonstrate that TokenTrim provides stronger and more consistent improvements than FlowMo across nearly all dimensions, particularly those related to long-horizon stability. Across the full set of VBench dimensions, TokenTrim achieves the best score in the majority of categories, whereas FlowMo only attains the highest score in a small subset and often trails both TokenTrim and the baseline.

Focusing first on motion-related metrics, TokenTrim yields substantial gains in \textbf{Temporal Flickering} (+2.12\% over baseline, from 96.77\% to 98.89\%) and \textbf{Motion Smoothness} (+1.81\%), outperforming FlowMo in both dimensions (95.21\% and 98.03\%, respectively). Importantly, TokenTrim also improves the \textbf{Dynamic Degree} (+2.54\%), while FlowMo significantly reduces it (-3.50\%), indicating that FlowMo’s apparent motion coherence may partially arise from suppressing motion complexity rather than preserving it. This contrast suggests that TokenTrim maintains realistic motion dynamics while reducing drift, rather than converging to overly static trajectories.

TokenTrim further improves higher-level semantic and structural consistency. It achieves the strongest gains in \textbf{Human Action} (+4.19\%), \textbf{Scene} (+2.98\%), and \textbf{Overall Consistency} (+3.45\%), consistently outperforming FlowMo, which provides smaller or inconsistent improvements in these categories. Notably, FlowMo only surpasses TokenTrim in \textbf{Subject Consistency} and \textbf{Spatial Relationship}, but these gains do not translate into improved aggregate performance.

In the aggregated metrics, TokenTrim clearly dominates. It improves the \textbf{Semantic Score} by +3.53\% and the \textbf{Quality Score} by +5.58\%, resulting in a \textbf{Final Score} of 79.67\% (+4.55\% over baseline). In contrast, FlowMo yields only marginal improvements, achieving a Final Score of 75.81\% (+0.69\%). These results indicate that while FlowMo offers limited local benefits, it does not effectively mitigate the accumulation of errors across auto-regressive steps. TokenTrim’s explicit control over context reuse enables significantly better long-horizon coherence and overall video quality.

In addition, a closer inspection of FlowMo’s per-dimension performance reveals that its improvements are uneven and largely confined to a small subset of metrics. Under Rolling Forcing, FlowMo provides modest gains in \textbf{Motion Smoothness} (+0.82\%, from 97.21\% to 98.03\%) and \textbf{Spatial Relationship} (+1.27\%), 
but these come at the expense of reduced motion diversity, as reflected by a substantial drop in 
\textbf{Dynamic Degree} (-3.50\%, from 59.57\% to 56.07\%). FlowMo also degrades several stability-sensitive 
metrics, including \textbf{Temporal Flickering} (-1.56\%) and \textbf{Overall Consistency} (from 24.92\% to 26.98\%, still well below TokenTrim’s 28.37\%). A similar pattern is observed under Self Forcing. While FlowMo yields a small improvement in \textbf{Quality Score} (+0.96\%), it reduces the \textbf{Semantic Score} (-0.73\%) and lowers the \textbf{Dynamic Degree} from 67.75\% to 65.46\%. Across both baselines, FlowMo attains the best score in only a few isolated dimensions (e.g., Subject Consistency and Spatial Relationship), but these gains do not translate into consistent improvements in long-horizon stability or aggregated performance.

\section{User Study: Instructions Provided to Participants}
\label{app:user_study}
As part of the evaluations we performed on our method, we conducted a user study, as described in \ref{token_trim_quantitative_results}. The study was designed to assess human preferences on videos generated with and without FlowMo, using the videoJAM benchmark~\cite{chefer2025videojam}, which focuses on motion coherence.

The study was conducted using Google Forms. For each prompt, participants were shown a pair of videos—one with FlowMo and one without—generated with the same random seed (1024). The order of the videos was randomized to avoid positional bias. Each pair was evaluated by five different participants, resulting in $640$ responses per baseline.

\begin{figure}[h]
  \centering
  \includegraphics[width=0.85\columnwidth]{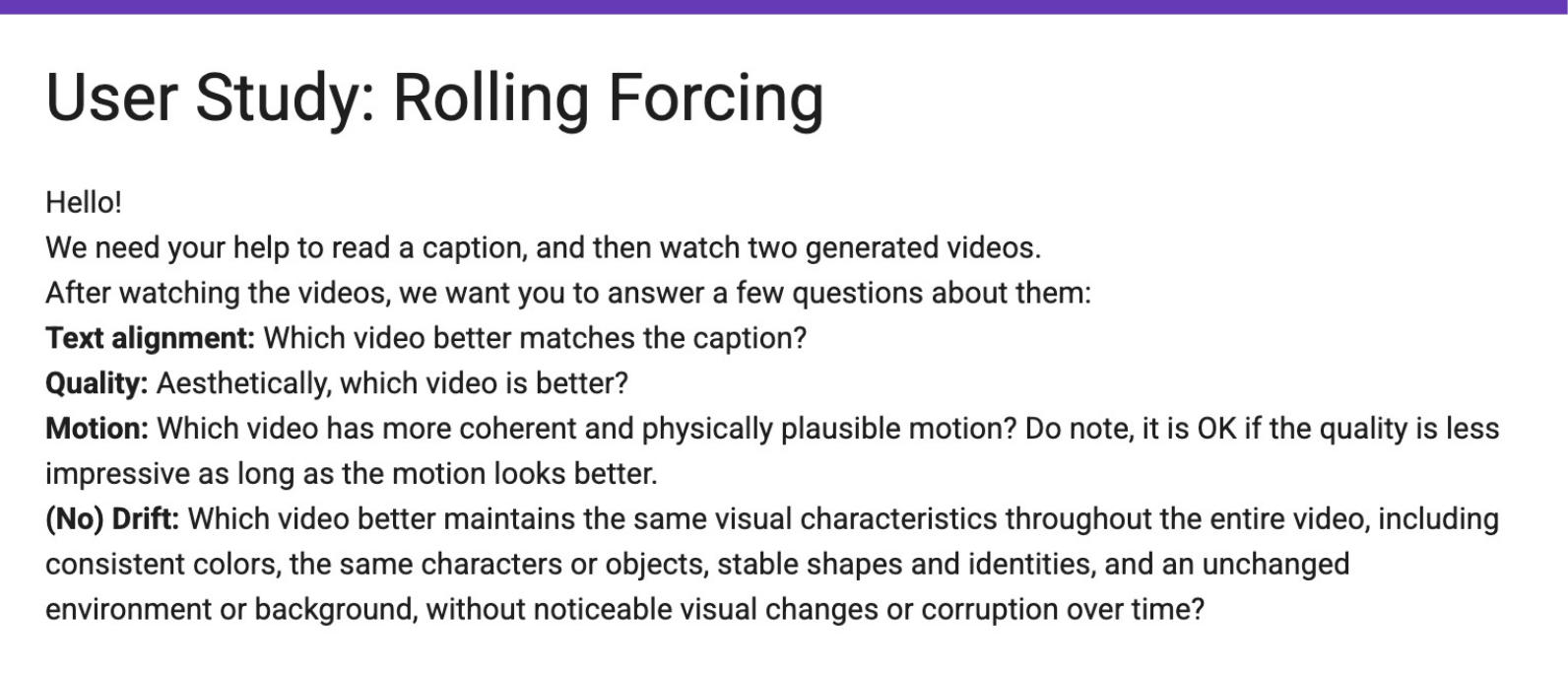}
  \caption{Screenshot of the Google Form used in the user study.}
  \label{fig:user_study_rolling_forcing}
\end{figure}

Participants were asked to evaluate the videos based on three criteria: \textbf{\textit{text alignment}}, \textbf{\textit{aesthetic quality}}, \textbf{\textit{motion coherence}}, and \textbf{\textit{(No) Drift}}. The instructions provided to the annotators are reproduced below, together with a screenshot of the annotation interface (see Fig.~\ref{fig:user_study_rolling_forcing}).

\paragraph{\textbf{Annotator Instructions.}}
Participants were first asked to carefully read the given text prompt and then watch two generated videos. After viewing both videos, they were instructed to answer the following questions:

\begin{itemize}
  \item \textbf{Text alignment:} Which video better matches the given caption?
  \item \textbf{Quality:} From an aesthetic perspective, which video looks better overall?
  \item \textbf{Motion:} Which video exhibits more coherent and physically plausible motion?
  \begin{itemize}
    \item \emph{Do Note: It is OK if the quality is less impressive as long as the motion looks better}
  \end{itemize}
  \item \textbf{(No) Drift:} Which video better maintains consistent visual characteristics throughout the entire sequence, including stable colors, characters or objects, shapes and identities, and an unchanged environment or background, without noticeable visual corruption or changes over time?
\end{itemize}

\section{Ablation Study - VBench Metrics Breakdown}
\label{app:ablation_study}
We present an ablation study in Tab.~\ref{tab:full_ablation} to analyze the contribution of each component of \textbf{TokenTrim} and to examine its sensitivity to pruning strength and initialization quality. All variants are evaluated on Rolling Forcing to isolate the effect of inference-time context management.

First, we observe that the full TokenTrim configuration consistently achieves the strongest performance across most motion- and stability-related dimensions, confirming that both adaptive pruning and drift-based triggering are essential. Compared to the baseline, TokenTrim improves \textbf{Temporal Flickering} by +2.12\%, \textbf{Motion Smoothness} by +1.81\%, \textbf{Dynamic Degree} by +2.54\%, and \textbf{Overall Consistency} by +3.45\%, indicating effective mitigation of long-horizon error accumulation without suppressing motion.

Next, we study the effect of pruning strength. A conservative pruning rate of \textbf{5\%} already yields noticeable improvements over the baseline, increasing the \textbf{Final Score} by +3.23\%. However, its gains are consistently smaller than those achieved by the full TokenTrim configuration, particularly in \textbf{Human Action} (+1.48\% vs. +4.19\%) and \textbf{Overall Consistency} (+0.73\% vs. +3.45\%). In contrast, aggressive pruning (\textbf{20\%}) leads to severe degradation across nearly all dimensions, including large drops in \textbf{Motion Smoothness} (-11.69\%), \textbf{Dynamic Degree} (-3.60\%), and \textbf{Quality Score} (-14.43\%), resulting in a substantial \textbf{Final Score} decrease of -11.87\%. This clearly demonstrates that excessive token removal harms semantic and visual fidelity, underscoring the necessity of adaptive, drift-aware pruning rather than fixed-rate pruning.

Finally, removing FlowMo from the initialization stage (\textbf{w/o FlowMo}) degrades performance relative to the full TokenTrim pipeline, reducing the \textbf{Final Score} by 2.22\%. While this variant still outperforms the baseline, it exhibits weaker gains in \textbf{Dynamic Degree}, \textbf{Human Action}, and \textbf{Overall Consistency}, highlighting the importance of a stable initial latent anchor for preventing early corruption that later propagates through the auto-regressive context.

Overall, this ablation study confirms that TokenTrim’s effectiveness arises from the combination of (i) drift-triggered pruning, (ii) moderate, adaptive pruning ratios, and (iii) stable initialization. Removing or weakening any of these components leads to measurable degradation, while overly aggressive pruning causes catastrophic loss of video quality.

\section{Additional Qualitative Experiments: TokenTrim vs Self Forcing}
\label{app:additional_qualitative_self}
\begin{figure*}[t]
  \centering
  \includegraphics[width=0.91\textwidth]{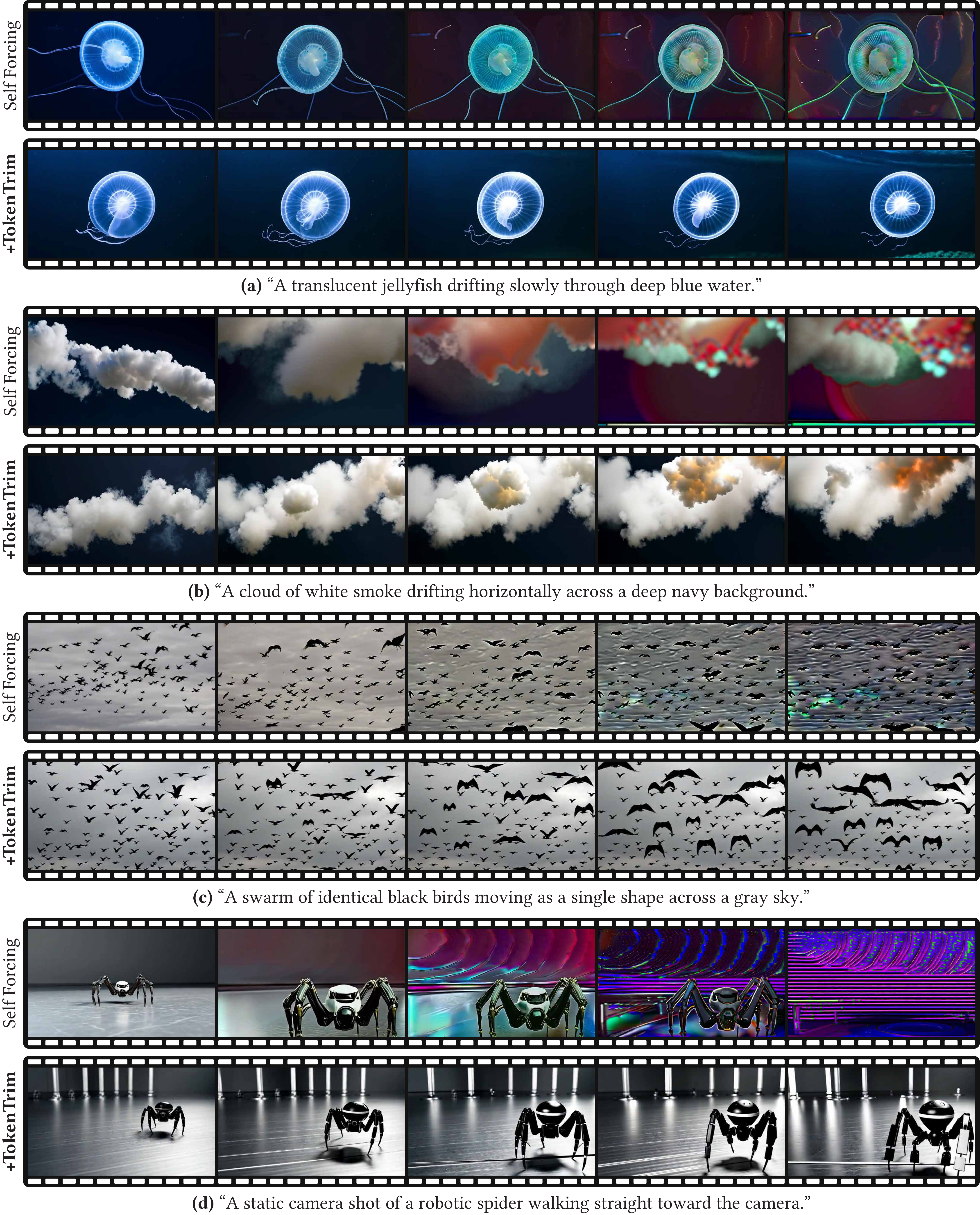}
  \caption{Additional Qualitative Results comparison between TokenTrim and Self Forcing~\cite{huang2025self}.}
  \label{fig:additional_qualitative_self}
\end{figure*}

Fig.~\ref{fig:additional_qualitative_self} compares TokenTrim with the Self Forcing baseline under identical long-horizon generation settings. Although Self Forcing improves short-term temporal anchoring, it still exhibits gradual degradation when errors accumulate across extended rollouts. In Fig.\ref{fig:additional_qualitative_self}(a), depicting a translucent jellyfish drifting through deep blue water, Self Forcing introduces subtle shape distortions and loss of translucency over time, while TokenTrim preserves a stable body structure and consistent appearance. In Fig.\ref{fig:additional_qualitative_self}(b), where a cloud of white smoke drifts horizontally across a dark background, Self Forcing suffers from spatial warping and uneven density, whereas TokenTrim maintains coherent global motion and uniform texture. Similar behavior is observed in Fig.\ref{fig:additional_qualitative_self}(c), where a swarm of birds generated with Self Forcing gradually loses collective structure, while TokenTrim preserves synchronized motion and stable silhouettes. Finally, in Fig.\ref{fig:additional_qualitative_self}(d), a robotic spider walking toward the camera exhibits accumulated structural drift under Self Forcing, whereas TokenTrim maintains consistent limb geometry and forward motion. Overall, these examples demonstrate that TokenTrim more effectively suppresses long-horizon error accumulation than Self Forcing, resulting in improved structural stability and temporal coherence across diverse motion patterns.

\section{Additional Qualitative Experiments: TokenTrim vs Rolling Forcing}
\label{app:additional_qualitative_rolling}

\begin{figure*}[t]
  \centering
  \includegraphics[width=0.91\textwidth]{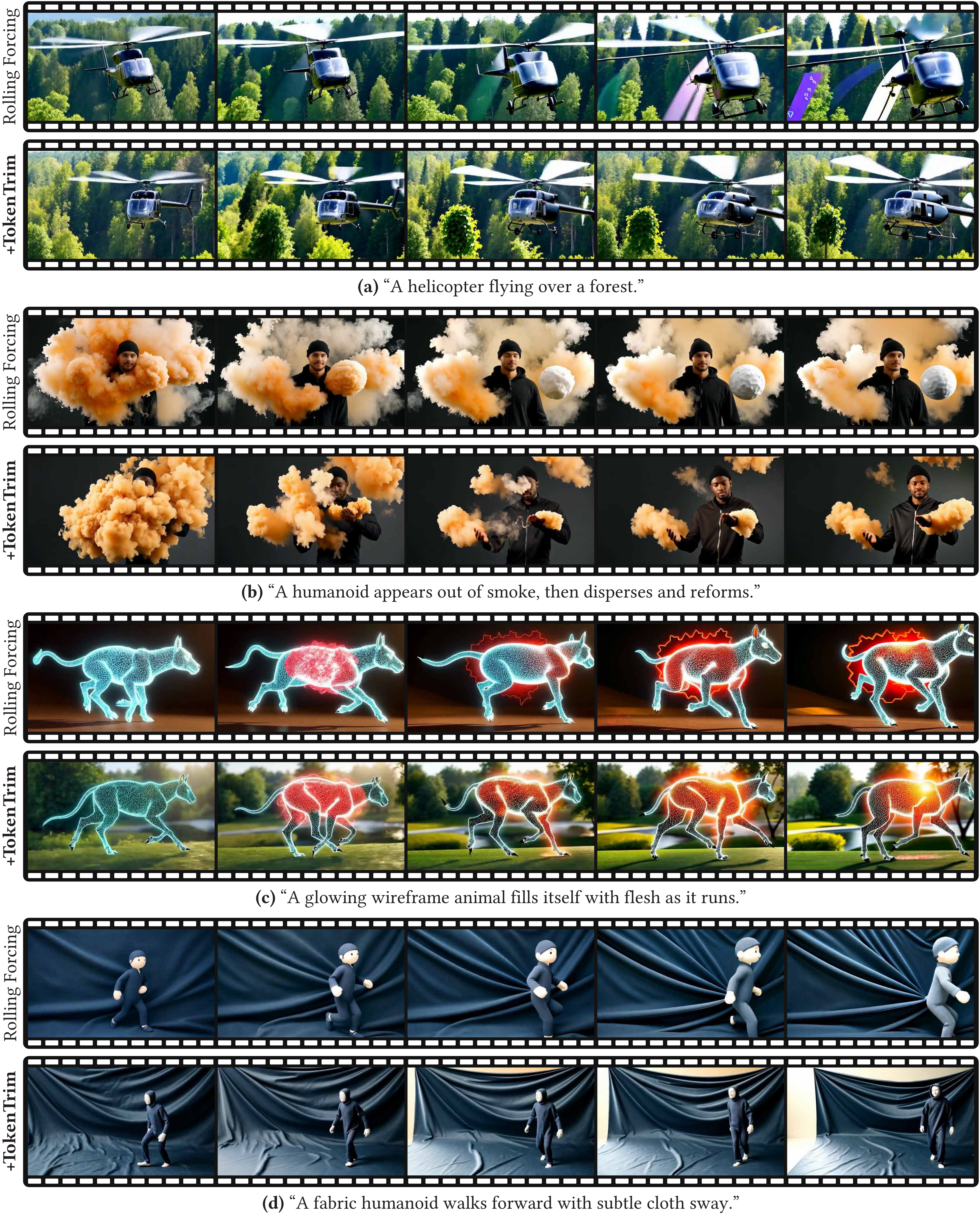}
  \caption{Additional Qualitative Results comparison between TokenTrim and Rolling Forcing~\cite{liu2025rolling}. }
  \label{fig:additional_qualitative_rolling}
\end{figure*}

Fig.~\ref{fig:additional_qualitative_rolling} compares long-horizon generations produced by Rolling Forcing with and without TokenTrim under identical settings. While Rolling Forcing generates plausible content in early frames, it exhibits progressive temporal degradation as the rollout advances. In Fig.\ref{fig:qualitative}(a), depicting a helicopter flying over a forest, the baseline gradually introduces background distortion and structural inconsistencies in the helicopter, whereas TokenTrim preserves stable geometry and consistent background appearance throughout the sequence. In Fig.\ref{fig:qualitative}(b), where a humanoid emerges from smoke and reforms, Rolling Forcing suffers from identity drift and shape instability during the reformation process, while TokenTrim maintains coherent structure and consistent appearance across frames. Fig.\ref{fig:qualitative}(c) shows a glowing wireframe animal filling itself with flesh while running: the baseline exhibits color bleeding and deformation over time, whereas TokenTrim preserves smooth transitions and stable anatomy. Finally, in Fig.\ref{fig:qualitative}(d), Rolling Forcing introduces cloth distortion and background warping during the walking motion, while TokenTrim maintains consistent cloth dynamics and a stable scene layout. Overall, these examples demonstrate that TokenTrim effectively suppresses long-horizon drift by preventing corrupted latent tokens from propagating through the auto-regressive context, resulting in significantly improved temporal stability compared to Rolling Forcing alone.

\section{Additional Qualitative Experiments: TokenTrim vs FlowMo}
\label{app:additional_qualitative_flowmo}

\begin{figure*}[t]
  \centering
  \includegraphics[width=0.91\textwidth]{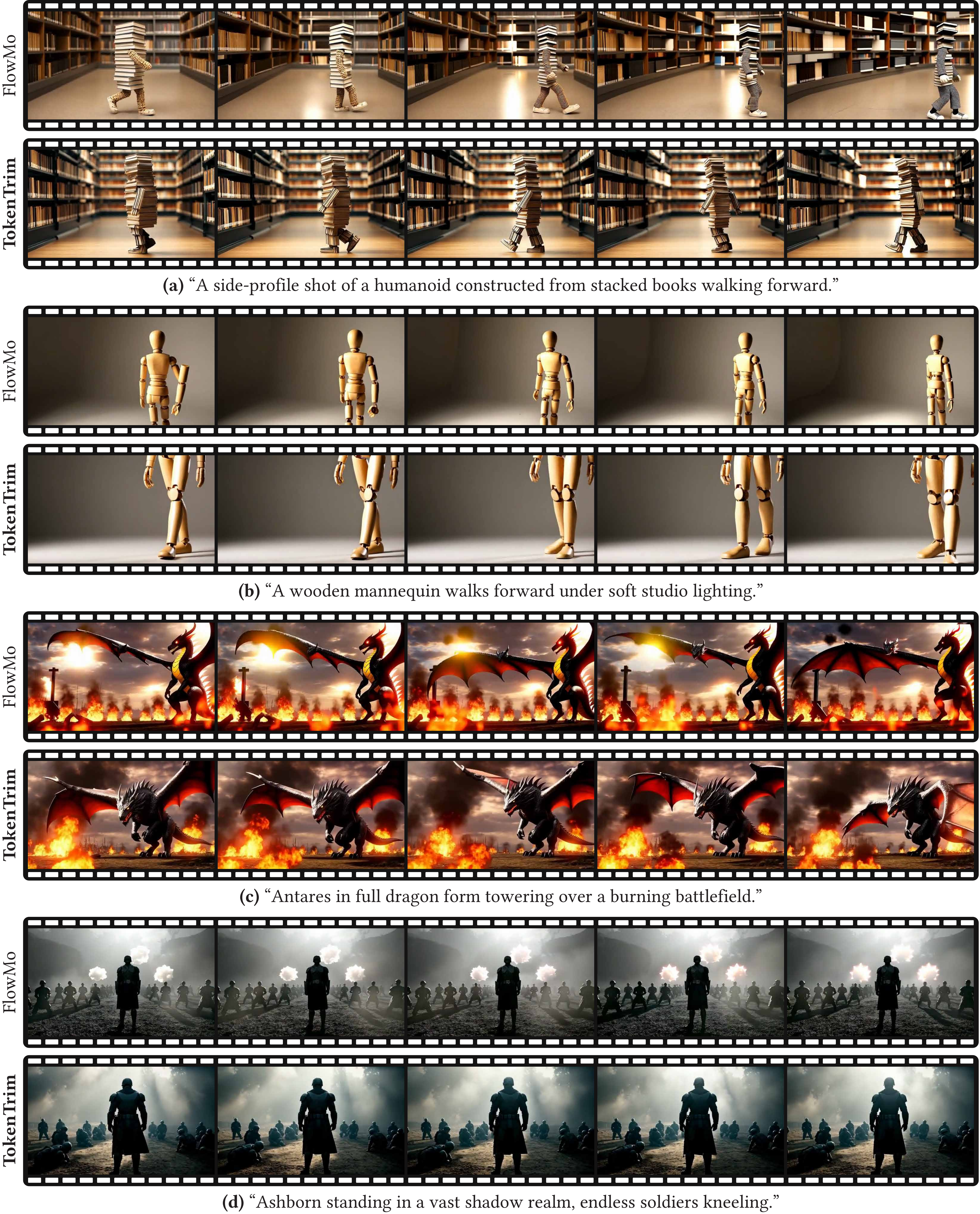}
  \caption{Additional Qualitative Results comparison between TokenTrim and FlowMo~\cite{shaulov2025flowmo}. }
  \label{fig:additional_qualitative_flowmo}
\end{figure*}

Fig.\ref{fig:additional_qualitative_flowmo} presents a qualitative comparison between TokenTrim and FlowMo under identical long-horizon generation settings. While FlowMo often produces plausible motion in early frames, it exhibits progressive structural and semantic drift as the rollout proceeds. In Fig.\ref{fig:additional_qualitative_flowmo}(a), showing a humanoid constructed from stacked books walking forward, FlowMo gradually introduces misalignment and deformation in the book stack, leading to inconsistent body proportions, whereas TokenTrim preserves a stable silhouette and coherent limb structure throughout the sequence. In Fig.\ref{fig:additional_qualitative_flowmo}(b), depicting a wooden mannequin walking under soft studio lighting, FlowMo suffers from subtle identity drift and joint instability over time, while TokenTrim maintains consistent articulation and appearance. Fig.\ref{fig:additional_qualitative_flowmo}(c) shows a dragon towering over a burning battlefield: FlowMo introduces background warping and deformation in the wings and body across frames, whereas TokenTrim preserves coherent anatomy and stable environmental structure. Finally, in Fig.~\ref{fig:additional_qualitative_flowmo}(d), FlowMo exhibits temporal inconsistency in the arrangement of background figures and lighting, while TokenTrim maintains a stable composition and consistent scene layout.

\end{document}